\documentclass[10pt,twocolumn,letterpaper]{article}

\usepackage{cvpr}
\usepackage{times}
\usepackage{epsfig}
\usepackage{graphicx}
\usepackage{amsmath}
\usepackage{amssymb}
\usepackage{multirow}

\usepackage[breaklinks=true,bookmarks=false]{hyperref}

\cvprfinalcopy 

\begin{document}

\title{Anomaly Detection and Localization in Crowded Scenes by Motion-field Shape Description and Similarity-based Statistical Learning}

\author{Xinfeng ZHANG$^{\dag}$, Su YANG$^{\dag}$\thanks{Corresponding author}, Xinjian ZHANG$^{\dag}$, Weishan ZHANG$^{\S}$, Jiulong ZHANG$^{\ddag}$ \\
  $^{\dag}$School of Computer Science, Fudan University\\
$^{\S}$Department of software engineering, China University of Petroleum\\
$^{\ddag}$School of Computer Science, Xi'an University of Tecgnology\\
{\tt\small \{10110240026, suyang, zhangxj17\}@fudan.edu.cn,{zhangws}@upc.edu.cn,{zjl}@xaut.edu.cn }
}

\maketitle

\begin{abstract}
   In crowded scenes, detection and localization of abnormal behaviors is challenging in that high-density people make object segmentation and tracking extremely difficult. We associate the optical flows of multiple frames to capture short-term trajectories and introduce the histogram-based shape descriptor referred to as shape contexts to describe such short-term trajectories. Furthermore, we propose a K-NN similarity-based statistical model to detect anomalies over time and space, which is an unsupervised one-class learning algorithm requiring no clustering nor any prior assumption. Firstly, we retrieve the K-NN samples from the training set in regard to the testing sample, and then use the similarities between every pair of the K-NN samples to construct a Gaussian model. Finally, the probabilities of the similarities from the testing sample to the K-NN samples under the Gaussian model are calculated in the form of a joint probability. Abnormal events can be detected by judging whether the joint probability is below predefined thresholds in terms of time and space, separately. Such a scheme can adapt to the whole scene, since the probability computed as such is not affected by motion distortions arising from perspective distortion. We conduct experiments on real-world surveillance videos, and the results demonstrate that the proposed method can reliably detect and locate the abnormal events in the video sequences, outperforming the state-of-the-art approaches.
\end{abstract}

\section{Introduction}

Due to the arising demand for public security issues, it is urgent to develop an automated system that can monitor and percept human activities to alarm abnormal events. In surveillance videos, the dominant activities occurring frequently are referred to as normal behaviors~\cite{6619181-17}. Apart from the normal activities, the most important and challenging task is to detect and localize anomalous events, which are defined as those to occur with a low probability~\cite{7054448-21}. In general, an abnormal event appears rarely and disappears in a short time~\cite{6247917-18}. The goal of anomaly detection and localization is to identify the small time span and the spatial region covering the anomalous activities in an automatic manner~\cite{5206771-11,5539872-12}.

In surveillance videos of crowded scenes, high-density people~\cite{5206771-11} make anomaly detection especially challenging. The traditional object-based approaches conduct anomaly detection based on objects’ appearances and trajectories, its performance is directly dependent on the accuracy of object extraction~\cite{1597122-13,Tu2008-22} and object tracking~\cite{4587510-3,Sethi:2010:MRC:1878039.1878049-19,Wang2006-23}. Unfortunately, capturing the single individuals is nearly impossible in crowded scenes, because of the high density of people and various objects performing irregular motions to incur frequent and severe occlusions~\cite{5206771-11}. In addition, tracking multiple objects is quite time-consuming~\cite{6844850-26}.

The latest trend in terms of anomaly detection is shifted to partition the surveillance videos into a couple of spatio-temporal volumes of a fixed size to focus on local scenes of a short time duration~\cite{7298909-9}. Then, the volume-based detection model in temporal and spatial contexts is established to discriminate whether the local scenes correspond with abnormal events or not, where the anomalies refer to such patterns that have never appeared at a specified site in contrast to the historical records or deviate remarkably from those of their neighborhoods at the same time~\cite{5539872-12}. In the literature, the unsupervised framework that uses normal volumes only for training has drawn considerable attention, since anomalies are always rare and differ from one to another with unpredictable variations, making it impossible to model all the abnormal types~\cite{BERTINI2012320-5,6619181-17}.

\begin{figure}[h]
  \centering
  \includegraphics[width=0.95\linewidth]{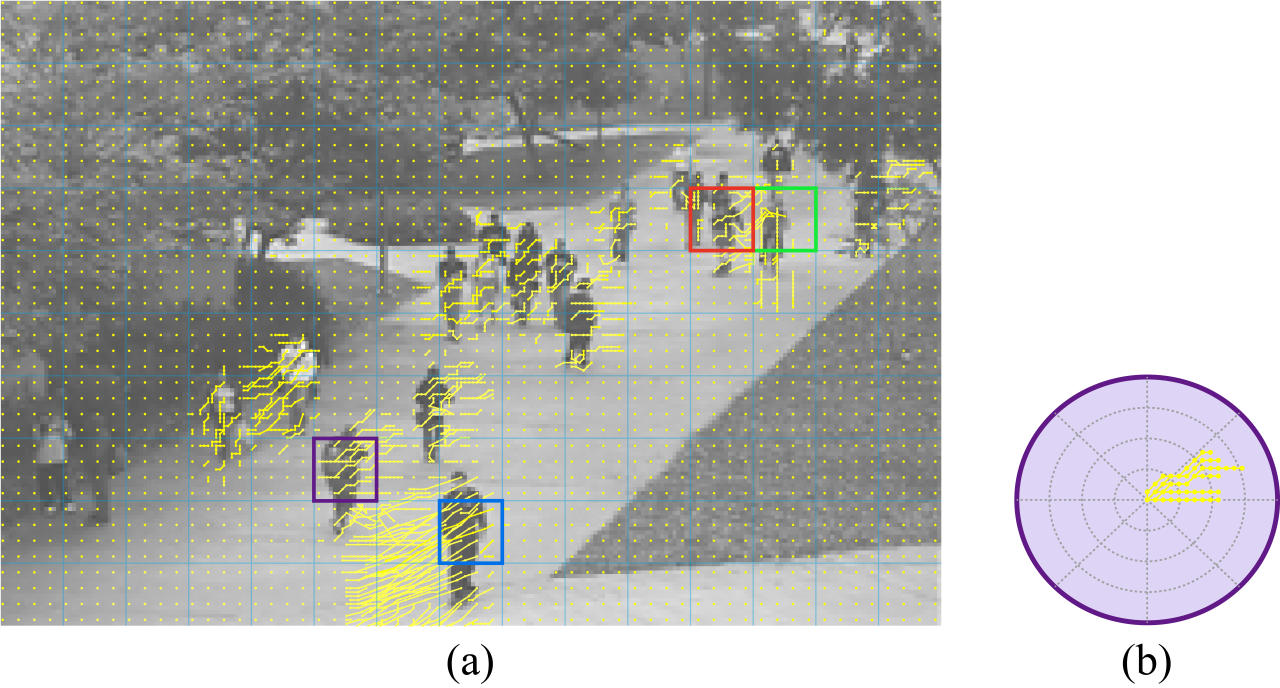}
  \caption{An example of short-term trajectories and the histogram-based shape description for trajectories. (a) The blocks with red, blue, and green colors correspond with anomalies, namely, some skaters and one biker, while the purple region is a normal case with pedestrian only. (b) The histogram to figure out the short-term trajectories in the purple block in (a).}
  \label{fig:1}
\end{figure}

There are two categories of unsupervised approaches for anomaly detection: (1) Apply clustering methods to find outliers~\cite{5539872-12}. In fact, such a scheme has been widely used in the existing works~\cite{BERTINI2012320-5,7298909-9,5206771-11,5539872-12,JAVANROSHTKHARI20131436-16,5539882-20}. However, how to determine the number of clusters remains unsolved. Classical clustering algorithms such as k-means and Gaussian Mixture Model (GMM)~\cite{7298909-9,5539872-12} require the number of prototypical patterns to be known a priori~\cite{5206771-11}. In crowded scene, however, motion patterns are changing continuously and randomly such that some of them cannot be foreseen, which leads to uncertainty in regard to the number of prototypical patterns. An alternative solution is to perform clustering based on a distance threshold so as to determine whether a sample belongs to an existing prototypical pattern or corresponds to a new prototype that should be created~\cite{5206771-11,JAVANROSHTKHARI20131436-16,5539882-20} as well as whether two clusters should be merged or not~\cite{BERTINI2012320-5}. For this kind of methods, a specific distance threshold applicable to the whole scene does not exist due to the size variation of the object of interest subject to the distance to the camera, namely, perspective distortion. For example, as shown in Fig.\ref{fig:1}(a), the size of the skater in red color is much smaller than that of the one in blue color. This causes the same problem in defining the number of prototypes.

(2) The other category of methods is reconstruction-based approaches. Yang \etal \cite{5995434-25} reconstruct testing samples from the normal samples of previous or surrounding volumes that act as the dictionary, and identify the samples with large reconstruction errors as anomalies. However, once a very small number of abnormal samples are mixed into the dictionary, it will fail to detect the same kinds of abnormal behaviors. Besides, it has the same problem in finding a threshold applicable to the whole scene on account of the perspective distortion.

Therefore, we propose a motion-field shape description along with a K-NN (K-nearest neighbors) similarity-based statistical model to detect anomalies over time and space, where clustering or prior assumption are not needed. First, we associate the optical flows across multiple frames to capture the short-term trajectories in a video clip. Hereafter, we introduce the histogram-based shape descriptor referred to as shape contexts~\cite{993558-4} to figure out the short-term trajectories within each patch in a statistical sense, which reflects faithfully the motion trend and details in every local patch. Then, we propose to compute the K-NN similarity-based statistical model as follows: First, we retrieve the K-NN samples from the training set in regard to the testing sample, and then use the similarities between every pair of the K-NN samples in the training set to construct a Gaussian model. Finally, the probabilities of the similarities from the testing sample to the K-NN samples under the Gaussian model are calculated in the form of a joint probability to check whether they are compatible with the Gaussian model. Abnormal events can be detected by judging whether the joint probability is below predefined thresholds in temporal and spatial contexts, separately. The major advantage is: The anomaly detection through probability can adapt to the whole scene, since the probability computed as such is not affected by the so-called perspective distortion. We carried out extensive experiments on two benchmarks with real-world scenes, UCSDped1 dataset~\cite{5539872-12} and Subway dataset~\cite{4407716-1}, for anomaly detection and localization, and the results validate the effectiveness of the proposed method. Moreover, its performance in localizing anomalies outperforms those of the state-of-the-art approaches.

\section{Related Work}

To overcome the difficulty in segmenting and tracking individual objects in crowded scenes, an emerging trend is to establish the detection model from local primitives such as pixels, image blocks/patches, and 3D cuboids/bricks~\cite{7298909-9}. Such local features can be sorted into 3 classes: (1) Local property based feature: Adam \etal \cite{4407716-1} use histograms of optical flows at specific regions to derive decision rules for anomaly detection. Mahadevan \etal \cite{5539872-12} employ a mixture of dynamic textures to describe jointly the appearances and the dynamics of local portions of videos in crowded scenes. This approach captures both temporal and spatial anomalies at the cost of highly intensive computations. Reddy \etal \cite{5981799-15} extract multiple features including motion, size, and texture from each foreground region and the decision resulting from each feature is fused to make the final decision. (2) The well-known interaction based feature is the social force model (SFM) introduced by Mehran \etal \cite{5206641-14}, where crowd actions are modeled as interaction forces estimated from the corresponding optical flow field. However, SFM is not reliable and robust enough in the present of disturbance. (3) Wu \etal \cite{5539882-20} make use of chaotic invariants as a trajectory feature, which is known as maximum Lyapunov exponent and correlation dimension, to measure how much neighboring particles deviate from their original closeness to each other after a certain steps of evolution. However, for the scenes that people’s movements are spatially constrained such as in corridors and underpasses, Lyapunov exponent may not reflect exactly the chaotic degree of the collective human mobility due to the constrained evolution of the trajectories. Furthermore, this feature is only applicable to the specific type of anomaly known as crowd chaos, which is the same as the entropy-based~\cite{Zhang2016-27} and energy-based features~\cite{XIONG2012121-24}.

Another important issue for anomaly detection is machine learning. The problem for supervised learning is that the annotations of normal and abnormal samples are difficult to be obtained since anomalies happen rarely and the diversity between each other with unpredictable variations makes modeling all the abnormal classes in advance impossible~\cite{BERTINI2012320-5,6619181-17}. Consequently, recent works favor unsupervised machine learning. According to different unsupervised models applied, we broadly classify the anomaly detection approaches into three categories:

(1) For clustering-based approaches, Roshtkhari and Levine~\cite{JAVANROSHTKHARI20131436-16} construct a codebook and calculate the probability of the spatio-temporal collection of the volumes according to the codebook, where predefining the largest number of Gaussians used in the GMM is required. Cheng \etal \cite{7298909-9} cluster the local features extracted around interesting points into a low-level visual vocabulary using the k-means algorithm and then measure the distances of a testing cuboid against the visual vocabulary to detect local anomalies. They cluster the collection of the features from nearby interesting points to build a high-level codebook of templates using a greedy clustering algorithm, and then construct a model for each template by fitting into a multivariate Gaussian distribution to detect global anomalies. However, using high-dimensional features to train a multivariate Gaussian distribution is subject to overfitting. Besides, a series of problems prevent this kind of approaches from being applied to broad-spectrum applications. For example, due to the uncertainty of the motions in crowded scenes, it is impracticable to define the number of prototypical patterns in advance~\cite{5206771-11} for classical clustering algorithms, such as k-means and GMM~\cite{7298909-9,5539872-12}. An alternative solution is to perform clustering based on a distance threshold, for instance, the greedy clustering algorithm~\cite{7298909-9}. For this kind of solutions, a specific distance threshold applicable to the whole scene does not exist due to perspective distortion.

(2) For reconstruction-based approaches, Yang \etal \cite{5995434-25} identify the observed samples with large sparse reconstruction errors that exceed a predefined threshold as anomalies. The dynamic sparse coding approach detects the volumes with large reconstruction errors in reference to the online updating dictionary learned from usual events as anomalies through a threshold~\cite{5995524-28}. However, once a very small number of abnormal samples are mixed into the dictionary, this kind of approaches will fail to detect the same kind of abnormal behaviors. Besides, a fixed threshold specified in advance also suffers from perspective distortion.

(3) Another category of approaches is based on modeling relationships among normal volumes. Kim and Grauman~\cite{5206569-10} utilize a space-time Markov random field (MRF) to detect abnormal activities in a video sequence, where each node in the MRF graph corresponds to a local region in the video frames and neighboring nodes in both space and time are associated with links. A MRF model is built for regular behaviors and the cases not compatible with the learned model are considered as anomalies. In crowded scenes, the neighboring volumes may be taken from different parts of the same object or different objects. Moreover, due to dynamic occlusions, such volumes are dynamically changing. These result in very complex and uncertain relationships among volumes that go very easily beyond trained prototypes.

In this study, We detect the correspondences among the optical flows across multiple frames to capture short-term trajectories and employ the histogram-based shape descriptor referred to as shape contexts~\cite{993558-4} to characterize such short-term trajectories. Then, we model the motion features using the proposed statistical modeling of the K-NN similarities to detect anomalies over time and space.

The proposed method gains advantage over the aforementioned methods in the following aspects: (1) The proposed model does not rely on assumptions about the appearances and states of unusual behaviors, so it is not necessary to model prototypical patterns, which are usually not known in advance. (2) The method is not sensitive to such case that a small number of abnormal samples corrupt the normal training set, since the proposed anomaly detector is based on statistics and it is highly possible that the majority of the K-NN samples are clean data without corruption. (3) The anomaly detection through probability can adapt to the whole scene, since the probability is not affected by motion distortions arising from perspective distortion. (4) Here, abnormal events are identified as those to appear with a low probability. The physical meaning of the probability threshold is so explicit that can be set easily without any prior knowledge. (5) Solved as joint probability-represented resemblances among samples, the overfitting problem caused by high dimensionality can be avoided and the proposed method is computationally efficient for real-time detection.

\section{Short-term Trajectory Feature}

Most of the existing motion-based approaches deploy optical flows~\cite{4407716-1,5206641-14}, which capture motions between two successive frames only but fail to associate motions over multiple frames. We associate the optical flows of multiple frames to capture short-term trajectories and employ a histogram-based shape descriptor, namely, shape contexts~\cite{993558-4}, to characterize such short-term trajectories.

\subsection{Short-term Trajectory}

A given video of a crowd scene is divided into a series of non-overlapping clips, and each clip consists of a couple of frames streaming over a short time. Here, each clip is represented by a matrix of $ W \times H \times T $ size, where $ W \times H $ denotes the image resolution of every frame (width by height) and $ T $ is the number of sequential frames. We apply the general optical flow algorithm~\cite{Brox2004-6} to obtain the motion vectors denoted as follows:

\begin{equation}
  \left\{( u_w^t , v_h^t )| w \in [1,W],h \in [1,H], t \in [1,T-1]\right\}
  \label{1}
\end{equation}
{\noindent where $ u $ represents the horizontal velocity and $v$ the vertical velocity~\cite{5539882-20}. We assume that the particles overlaying on pixels move with the optical flows to form particle trajectories in a video clip~\cite{4270002-2}. The position of a moving particle is formulated as follows:}

\begin{equation}
\left\{\begin{aligned}
    x_w^{t+1} = x_t^w + [u_w^t]\\
    y_h^{t+1} = y_h^t + [v_h^t]
\end{aligned}
\right.
\label{2}
\end{equation}
{\noindent where $[\cdot]$ denote rounding operation and vector $(x_w^{t},y_h^{t})$ the position of particle $(w,h)$ at time $t$ . Following~\cite{5539882-20}, a particle trajectory is denoted as $\big\{(x_w^t , y_h^t) \vee t \in [1,T]\big\}$, and all the particle trajectories in a clip are denoted as:}

\begin{equation}
  \left\{( x_w^t , y_h^t )| w \in [1,W],h \in [1,H], t \in [1,T]\right\}
  \label{3}
\end{equation}
As an example illustrated in Fig.\ref{fig:1}(a), the yellow dot lines denote the short-term particle trajectories in a clip. Note that the positions of the particles are re-initialized for each clip, enabling the short-term trajectories to record only motions within each clip.

\subsection{Shape Histogram for Short-term Trajectory}

We divide the starting frame of a clip into non-overlapping small patches $\big\{ s(m) | m \in [1,M]\big\}$, where $ M $ denotes the total number of the patches and the frame partition should meet the condition that each patch does not involve too many objects to avoid interference with each other. Then, the short-term trajectories staring in the same patch $\big\{ ( x_w^t , y_h^t ) | (w , h) , t \in [1 , T]\big\}$ will undergo the histogram-based shape description as follows to characterize the corresponding motion patterns.

Firstly, perform translating on the particle trajectories in each patch to make the starting point of every trajectory locate at the origin of the polar coordinate, that is, $\big\{(x_w^t - x_w^1, y_h^t - y_h^1 ) | (w , h) \in s(m) , t \in [1 , T] \big\}$.

Then, arrange the particle trajectories into $ b_M \times b_A $ bins that are uniformly partitioned in terms of both magnitude and angle in the polar space, where $ b_M $ represents the number of the magnitude intervals and $ b_A $ that of the angle intervals. Finally, count the non-overlapping particles falling into each bin to obtain a histogram $\big\{ h(n) | n \in [1 , N] \big\}$, where $ h(n) $ denotes the number of the particles falling in the $ n^{th} $ bin and $ N = b_M \times b_A $ the total number of the bins. As shown in Fig.\ref{fig:1}(a), the starting point of every trajectory in the patch labeled with purple color is translated to the origin of the histogram in Fig.\ref{fig:1}(b) in order to figure out the distribution of the particles falling into each bin of the histogram.

In contrast to the trajectory feature of the chaotic invariants~\cite{5539882-20}, we apply the shape histogram, namely shape contexts~\cite{993558-4}, to describe short-term particle trajectories, which preserves the details and the trends of local motions. Even in case the movements are constrained in a narrow space, abnormal human mobility in terms of speed can also be reflected by the length of the trajectories in the shape histogram. Moreover, the histogram of short-term trajectories can also be spited into two histograms by accumulating the number of particles along the magnitude and angle dimension, respectively, which enable anomaly detection on speed and direction to be independent.

\section{Statistical Modeling of K-NN Similarities}

First, we use $\chi^2$ test to measure the similarity between a testing sample and the normal training data, and retrieve the K-NN samples from the given training set in regard to the testing sample. Then, we establish a Gaussian model for the K retrieved samples to characterize the similarities among them in a statistical sense by computing the similarities from every pair of the K-NN samples, which can be regarded as statistical modeling of the intra-class similarities in the local manifold of the normal samples retrieved by the input testing sample. Here, the K retrieved samples are modeled through a one-dimensional Gaussian model over similarity in order to overcome the overfitting problem. Note that even if the samples to be compared in terms of similarity are high-dimensional data, the similarity itself is one-dimensional metric.

\subsection{Gaussian Distribution over K-NN Similarities}

Let $ H_p = \big\{ h_p(n) | n \in [1 , N] \big\}$ and $ H_q = \big\{ h_q(n) | n \in [1 , N] \big\}$ denote the shape contexts based motion features of two patches, respectively. Following the definition in~\cite{993558-4}, we use the $ \chi^2 $ test to measure the similarity between the two histograms $ H_p $ and $ H_q $, that is,

\begin{equation}
  \chi^2 \big( H_p , H_q \big) = \frac{1}{2} \cdot \sum_{n=1}^{N} \frac{[ h_p(n) - h_q(n)]^2}{ h_p(n) + h_q(n)}
  \label{4}
\end{equation}

From training set, we retrieve the K patches whose motion patterns are similar at most to that of the testing sample $ H_T $ in the sense of Eq.(\ref{4}), which are denoted as $\big\{ H_{NN}(k) | k \in [1 , K] \big\}$. Then, we calculate the similarities between every pair of them to obtain $\big\{ \chi^2 (H_{NN}(i),H_{NN}(j))| i \neq j , i \in [1 , K] , j \in [1 , K]\big\}$. We model these intra-class similarities as a one-dimensional probabilistic model, that is, K-NN similarities-rendered Gaussian model $ N(\mu , \sigma^2) $, where $\mu$ and $\sigma^{2}$ denote the mean and variance of the similarities $\big\{ \chi^2 (H_{NN}(i),H_{NN}(j))| i \neq j , i \in [1 , K] , j \in [1 , K]\big\}$ with the definitions as follows:

\begin{equation}
\resizebox{.9\hsize}{!}{$\left\{\begin{aligned}
    &\mu = \frac{2}{K \cdot (K-1)} \sum_{i=1}^{K-1} \sum_{j = i+1}^{K} \chi^2(H_{NN}(i) , H_{NN}(j)) \\
    &\sigma^2 = \frac{2}{K \cdot (K-1)} \sum_{i=1}^{K-1} \sum_{j = i+1}^{K}[\chi^2(H_{NN}(i) , H_{NN}(j))-\mu]^2
\end{aligned}
\right.
  \label{5}$}
\end{equation}

\subsection{Similarity-rendered Joint Posterior Probability}

Once the similarities between the testing patch $H_T$ nd its K-NN patches $\big\{ H_{NN}(k) | k \in [1 , K] \big\}$ are obtained as $\big\{ \chi^2 (H_T,H_{NN}(k)) | k \in [1 , K] \big\}$, the fitness of $H_T$  into the K-NN similarities-rendered Gaussian distribution $N(\mu, \sigma^2)$ is calculate as the joint posterior probability $L_T$ defined below:

\begin{equation}
      L_T = \sum_{k =1}^{K}\log\big\{ Pr[ \chi^2 (H_T, H_{NN}(k)) \in N(\mu,\sigma^2)]\big\}
      \label{6}
\end{equation}
{\noindent where $\mbox{Pr}$ denote probability and the definition of the K-NN similarities-rendered Gaussian model $N(\mu,\sigma^2)$ refer to Eq.\ref{5}. The joint posterior probability tends to be 0 as the number of K-NN patches increases, so we compute the sum of the logarithms of the probabilities instead. Subsequently, we judge whether the testing sample is abnormal by comparing its joint probability with a user-defined threshold $T_p$. If $L_T < T_p$, as a low-probability event, the corresponding testing sample is classified as anomaly.}

The K-NN similarity-based statistical model for anomaly detection is reasonable in that: (1) For a normal testing patch, this patch and its K-NN patches are from the same normal class, so the similarities between this testing patch and its K-NN patches should follow the same distribution $N(\mu,\sigma^2)$. (2) For an abnormal testing patch, the similarities between it and the retrieved K-NN samples in the training data can be regarded as between-class similarities. In such a case, their similarities deviate from the intra-class similarities-rendered Gaussian distribution $N(\mu,\sigma^2)$ such that the possibility to follow the same distribution $N(\mu,\sigma^2)$ is low. Note that the proposed K-NN similarity-based statistical model does not rely on any specific feature, so that it is a generic machine learning method applicable to a variety of features for anomaly detection.

\section{Spatio-temporal Anomaly Detection}

For temporal anomaly detection at a given location, the training data are composed of the patches from the same location of a long history. For spatial anomaly detection, the training samples are from the surrounding patches. Two independent thresholds are used to detect the patches occurring with small probabilities in contrast to the temporal and spatial contexts, respectively.

For the scenes containing active regions and non-active regions that are less visited, a lower probability threshold is usually set for non-active regions than that for active regions, since the anomalies appear with obviously lower probabilities in the rarely visited non-active regions.
In crowded scenes, motion clues such as optical flows are not stable in that one object often falls into different patches and the detected patches are in general only a part of the entire abnormal objects, so we will extend the detected abnormal patches to the surrounding regions to cover the most part of abnormal objects. Here, we apply multi-scale analysis with two thresholds, a lower one to location the anomalies with high certainty and a higher one to spread the detected areas to enclose the major portions of the objects corresponding with anomalies.

\section{Experiments}

The proposed method is tested on public real-world datasets: The UCSDped1 dataset~\cite{5539872-12} and the Subway dataset~\cite{4407716-1} with varying densities of people. The challenge is that the scenes in the datasets are not only crowded but also with some extent of perspective distortion.

\subsection{Evaluation Criteria}

We evaluate different methods following the criteria widely used in previous works~\cite{BERTINI2012320-5,5206569-10,5539872-12,6619181-17}, which are as follows:

{\bfseries Frame-level evaluation}: If any region in a frame is identified as anomaly to be consistent with the ground truth, such detection is granted to be correct regardless of the location and the size of the region.

{\bfseries Pixel-level evaluation}: If over $ 40 \% $ portion of the ground truth are detected as anomalies in a frame, such detection is regarded as a right detection. So, pixel-level evaluation is stricter than frame-level evaluation.

As for quantitative evaluation, receiver operating characteristic ({\bfseries ROC}) reflects the relationship of true positive rate (TPR) against false positive rate (FPR), which are defined below:

{\bfseries True positive rate}: The rate of correctly detected frames to all abnormal frames in ground truth.

\begin{equation}
  \mbox{TPR} = \frac{\mbox{\# True Detection}}{\mbox{\# Abnormal Frames}}
  \label{7}
\end{equation}

{\bfseries False positive rate}: The rate of incorrectly detected frames to all normal frames in ground truth.
\begin{equation}
  \mbox{FPR} = \frac{\mbox{\# False Detection}}{\mbox{\# Normal Frames}}
  \label{8}
\end{equation}

ROC curve is plotted according to the detection results under different parameters. We quantify the performance in terms of the equal error rate ({\bfseries EER}) and the area under ROC curve ({\bfseries AUC}). The EER is the point on the ROC curve that FPR is equal to $\big( 1 - \mbox{TPR}\big)$. A smaller EER corresponds with better performance. As for the AUC, a bigger value corresponds with better performance.

{\bfseries Event-level evaluation}: If any position with true anomaly is detected and localized as abnormal, the detection is regarded as a correct hit~\cite{CONG20131851-7,5206569-10}.  On the other hand, if any normal frame is detected as anomaly, it is counted as a false alarm in terms of event detection~\cite{5995524-28}.

\subsection{Anomaly Detection and Localization on UCSDped1 dataset}

We conduct anomaly detection and localization test on the UCSDped1 dataset~\cite{5539872-12}, which records a large number of pedestrians walking on the walkway in a college campus to approach or move far away from the camera. Since the number as well as the density of the people appearing in the monitoring area varies with time and the video sequences are captured with a certain degree of perspective distortion, this benchmark is quite challenging. There are $34$ training video sequences of normal cases and $36$ testing video sequences involving various abnormal events, which include non-pedestrians on the walkway (\eg bikers, skaters, small carts, and people in wheelchairs etc.), as well as pedestrians with anomalous motion patterns or in non-walkway regions (\eg people running or walking across the grass etc.). Each video sequence has 200 frames with $158 \times 238$ resolution.

\begin{figure}[htbp]
  \centering
  \includegraphics[width=0.95\linewidth]{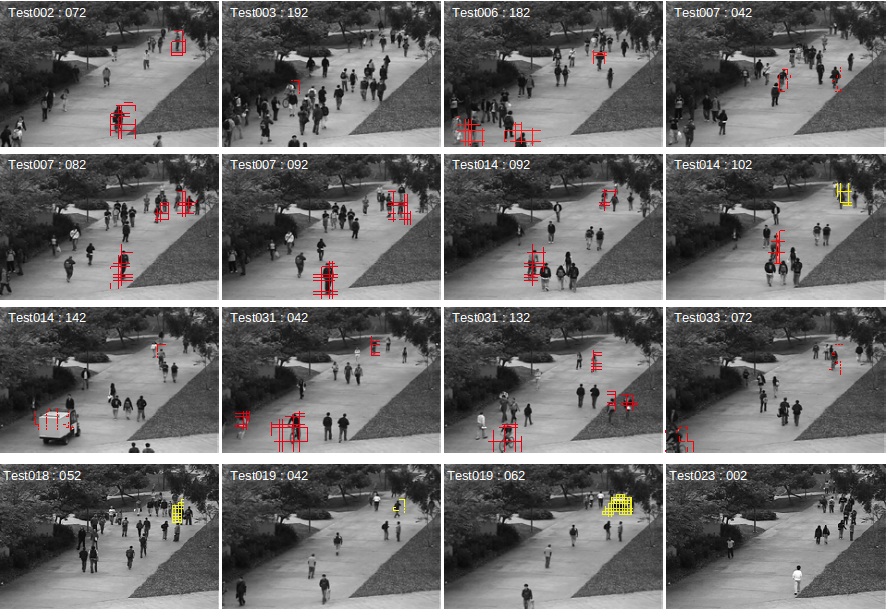}
  \caption{Some examples of the detection results on the UCSDped1 dataset. The red color means correct detection and the yellow color denotes false alarm according to the ground truth. The last row shows some true anomalies missing annotations in the ground truth but captured by the proposed method, where a pedestrian crossing the walkway in abnormal direction, a skate, the frontal of a vehicle, and a person on wheelchair take places in the frames labeled “Test019 : 042”, “Test018 : 052”, “Test019 : 062”, and “Test023 : 002”, respectively.}
  \label{fig:2}
\end{figure}

{\bfseries Results}: To enable detection of anomalies of small sizes, we uniformly divide the starting frame of every video clip into a couple of patches of $3 \times 3 \mbox{ pixels}$ without overlap, and then link the optical flows in the 10 consecutive frames of each video clip to obtain short-term trajectories. Some examples of the detection results achieved by using the proposed method are shown in Fig.\ref{fig:2}, which should be the most difficult tasks in the literature. One challenging issue is that multiple objects may appear in one scene. For instance, 4 persons with abnormal behaviors appear in the frames labeled “Test007 : 082” and “Test031 : 132”. Another difficulty is the occlusions caused by the high density of people. For example, a large portion of the biker in the frame labeled “Test003 : 192” is blocked and thus invisible due to the high density of people. In the aforementioned examples, the anomalies are detected correctly by using the proposed method, which are marked with red grids as shown in Fig.\ref{fig:2}. In Fig.\ref{fig:2}, we also illustrate some examples of false alarm, which are labeled in yellow color. To our surprise, some regions labeled in yellow color are indeed true anomalies such as the yellow region in the frame labeled “Test014 : 102”, which corresponds with a partially visible biker blocked by the vegetation and pedestrians. In the previous frame labeled “Test014 : 092”, the corresponding object is a true anomaly but it is incorrectly annotated in the ground truth~\cite{5539872-12}. The results manifest that the proposed method is able to detect different types of anomalies and performs well on very small patches. In another words, the proposed method can adapt to the whole scene with perspective distortion as it can detect the anomalies wherever they are, close to or far from the camera.

\begin{figure*}[htbp]
  \centering
  \includegraphics[width=0.8\linewidth]{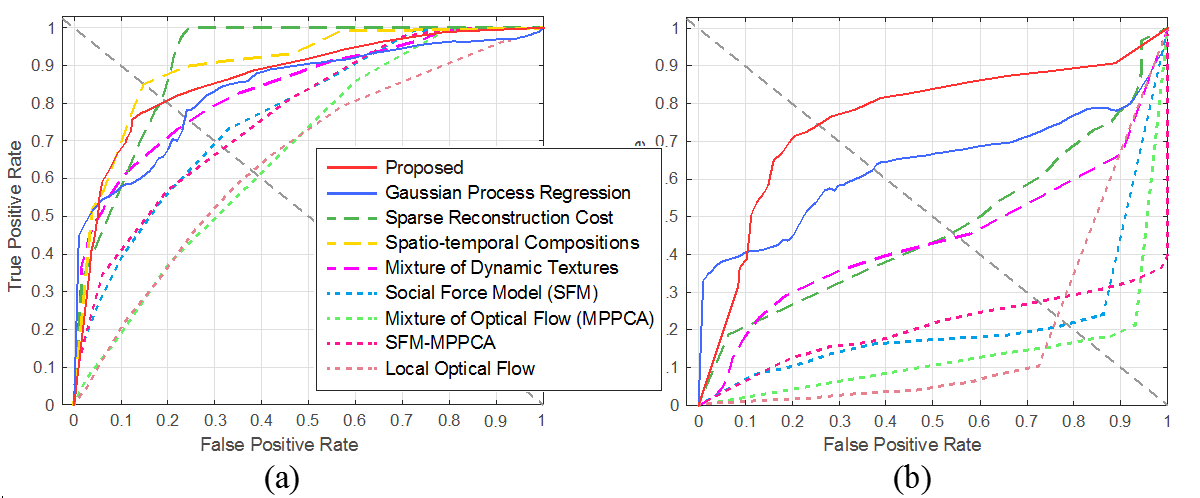}
  \caption{The comparison of ROC curves on the UCSDped1 dataset using different approaches. The dashed diagonal is the EER line. (a) Frame-level evaluation. (b) Pixel-level evaluation.}
  \label{fig:3}
\end{figure*}

{\bfseries Comparison with the existing approaches}: In Fig.\ref{fig:3}, we compare the ROC curves of the proposed method with Gaussian process regression (GPR)~\cite{7298909-9}, sparse reconstruction cost (SRC)~\cite{5995434-25}, spatio-temporal compositions (STC)~\cite{JAVANROSHTKHARI20131436-16}, mixture of dynamic textures (MDT)~\cite{5539872-12}, social force model (SFM)~\cite{5206641-14}, mixture of optical flow (MPPCA)~\cite{5206569-10}, SFM-MPPCA~\cite{5539872-12}, and local optical flow (LOF)~\cite{4407716-1}. In order to quantitatively evaluate the performances of different approaches, EER and AUC of anomaly detection at both pixel level and frame level as suggested by~\cite{5539872-12} are listed in Table \ref{Tab1}. For the frame-level evaluation, spatio-temporal compositions and sparse reconstruction cost achieve comparative performances in contrast to the proposed method.

\begin{table}[h]
  \caption{Comparison of EER and AUC using different methods}
\resizebox{0.95\linewidth}{!}{
  \begin{tabular}{|c|cccc|}
    \hline
    \multirow{2}{*}{Method} & \multicolumn{2}{|c|}{Frame Level} & \multicolumn{2}{|c|}{Pixel Level}  \\
    \cline{2-5}
     & \multicolumn{1}{|c}{ AUC(\%)} & EER(\%) & AUC(\%) & EER(\%) \\
    \hline
    \hline
    Proposed & 86.9 & 19.5 & 76.2 & {\bfseries25.6} \\
    GPR\cite{7298909-9} & 83.8 & 23.7 & 63.3 & 37.3 \\
    SRC\cite{5995434-25} & - & 19 & - & 54 \\
    STC\cite{JAVANROSHTKHARI20131436-16} & - & {\bfseries 15} & - & 27 \\
    MDT\cite{5539872-12} & - & 25 & - & 58 \\
    SFM\cite{5206641-14} & - & 31 & - & 78 \\
    MPPCA\cite{5206569-10} & - & 40 & - & 83 \\
    SFM-MPPCA~\cite{5539872-12}& - & 32 & - & 73 \\
    LOF\cite{4407716-1} & - & 38 & - & 76 \\
    \hline
    \multicolumn{5}{l}{AUC is the area under ROC curve;} \\
    \multicolumn{5}{l}{EER is equal error rate;- means the results are not 	provided} \\
  \end{tabular}
  }
  \label{Tab1}
\end{table}

However, frame-level evaluation does not consider whether the detection coincides with the actual location of the anomaly~\cite{5539872-12}. In contrast, pixel-level evaluation emphasizes the localization ability of an algorithm~\cite{JAVANROSHTKHARI20131436-16}. In surveillance videos, perspective distortion causes the motion vectors not consistent with each other to fall in diverse directions and scales. This will significantly affect anomaly detection in local regions. Since almost all approaches~\cite{7298909-9,JAVANROSHTKHARI20131436-16,5995434-25} treat all the volumes in a scene equally by applying uniform threshold or vocabulary without taking into account the aforementioned effect caused by perspective distortion, their performances degrade in locating anomalies. Since the probability threshold is capable of adapting to perspective distortion, the proposed method performs much better at pixel level as shown in Fig.\ref{fig:3}(b) and Table~\ref{Tab1}.

\subsection{Computational Complexity}

We compare the computational time of the proposed method with those of Gaussian process regression~\cite{7298909-9}, sparse reconstruction cost~\cite{5995434-25}, and spatio-temporal compositions~\cite{JAVANROSHTKHARI20131436-16} on the UCSDped1 dataset. The results are listed in Table~\ref{Tab2}. The proposed method is implemented using MATLAB and the experiments are performed on a computer with Core i7-2600 CPU and 16GB RAM. Our method requires less learning and inferring time, approximately 175 milliseconds per frame, which contains the shape feature extraction and the construction and inference of the K-NN similarity-based model. It is obviously that the proposed method is very efficient.

\begin{table}[h]
  \caption{Computational time of different approaches (processing time per frame in milliseconds)}
  \resizebox{0.95\linewidth}{!}{
  \begin{tabular}{|c|cc|}
    \hline
    Method & learning & Inferring \\
    \hline
    \hline
    Proposed &\multicolumn{2}{|c|}{total:175} \\
    Gaussian Process Regression$^\alpha$ & 140.2 & 515.3 \\
    Spatio-temporal Compositions$^\alpha$ & 2432.5 & 2424.1 \\
    Sparse Reconstruction Cost$^b$ & - & 3800 \\
    \hline
    \multicolumn{3}{l}{“-” means the results are not provided; $^\alpha$ The approximate}\\
    \multicolumn{3}{l}{ computational time are from~\cite{7298909-9} on computer with Core}\\
    \multicolumn{3}{l}{ i7-2600 CPU and 4GM RAM; $^b$ The running time on a  }\\
	\multicolumn{3}{l}{computer with 2.6GHz CPU and 2GB RAM taken from~\cite{5995434-25}.}
\end{tabular}}
\label{Tab2}
\end{table}

\subsection{Influence of K}

\begin{figure}[htbp]
  \centering
  \includegraphics[width=0.95\linewidth]{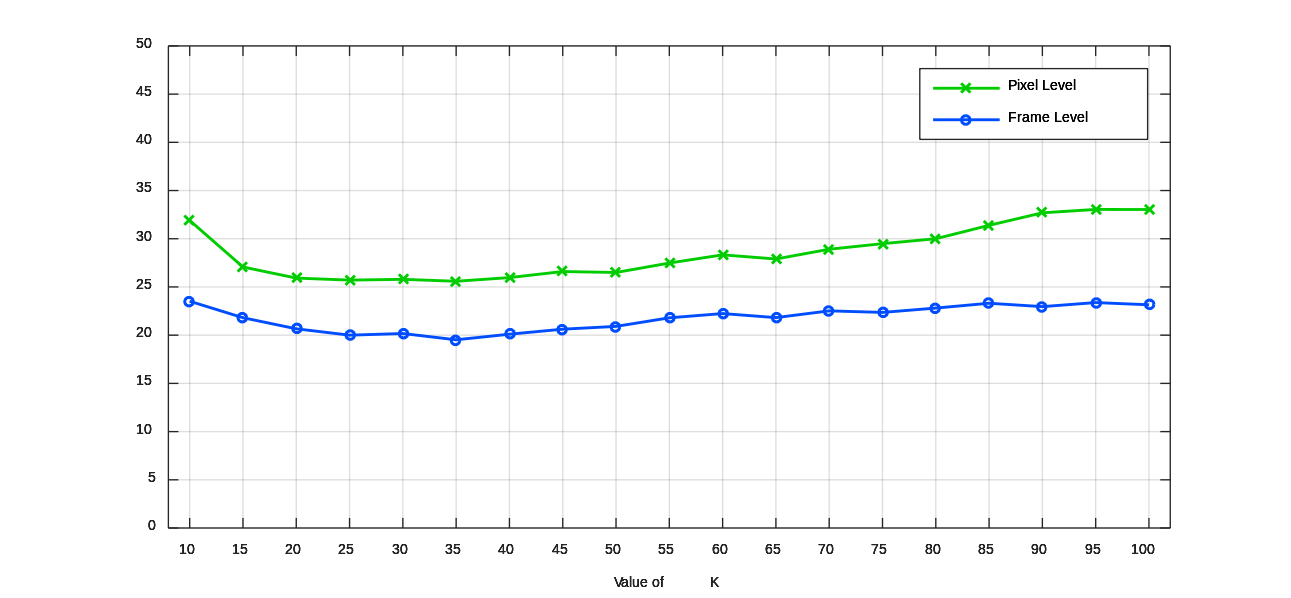}
  \caption{Influence of the value of K for anomaly detection on the UCSDped1 dataset}
  \label{fig:4}
\end{figure}

As K training samples similar to the testing sample at most are retrieved to establish the K-NN similarity-based Gaussian model, it is necessary to analyze the influence of K. EER for the UCSDped1 dataset at frame level and pixel level with different K are plotted in Fig.\ref{fig:4}. It can be seen that the EER varies little for different settings of the value of K from 15 to 70. This demonstrates that the proposed method is not sensitive to the value of K. In contrast, for the clustering algorithm, the number of the prototypical patterns has a strong impact on the clustering result.

\subsection{Evaluation on Subway dataset}

The Subway dataset is provided by Adam \etal \cite{4407716-1}. In the experiment, we use the surveillance videos at the exit of subway to examine the performances of different approaches. The camera is pointed toward the exit gate, where the dominant behaviors are exiting from the platform, coming up through the turnstiles, and turning to the left or right. The video lasts about 43 minutes with a resolution of $ 384 \times 512 \mbox{ pixels} $, which contains 19 anomalous events, mainly involving walking in the wrong direction, loitering, and miscellany~\cite{5206569-10}.

\begin{figure}[htbp]
  \centering
  \includegraphics[width=0.95\linewidth]{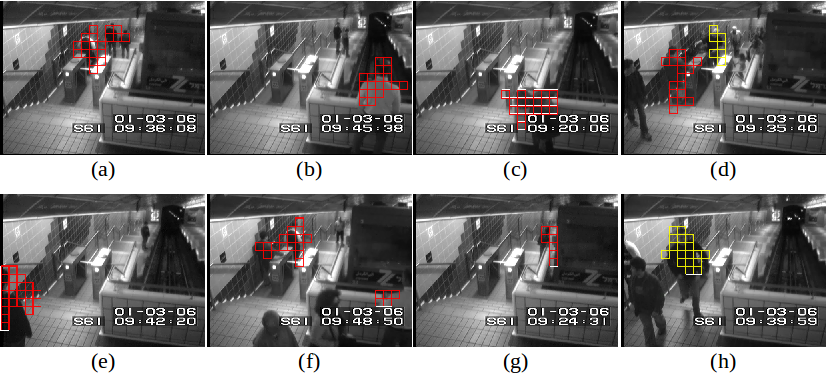}
  \caption{Examples of the detection results on the Subway dataset. The red color indicates correct detection and the yellow color denotes false alarm. (a) and (e) Wrong direction: Some persons are entering through the exit gate. (b) and (f) Loitering: A person is wandering; Two persons are entering through the exit gate. (c) and (g) Miscellany: A person is cleaning the wall; A person gets off the train and then gets on the train again very soon. (d) and (h) False alarm: An adult is helping a child pass the turnstile; A person is coming up from the turnstile with an irregular jump; A correct detection: A person is entering through the exit gate. }
  \label{fig:5}
\end{figure}

{\bfseries Result}: The starting frame of each clip is divided into non-overlapping patches of $20 \times20 \mbox{ pixels}$, and the length of each video clip is set to be 10 consecutive frames (400 milliseconds), where no overlap exists between two continuous clips. The first 10 minutes of the video are used for training, while the rest of the frames are used for testing. Some examples of the detection results achieved by using the proposed method are shown in Fig.\ref{fig:5}, where correct detections and false alarms are both included. The results validate that the proposed method with a fixed threshold can capture multiple abnormal objects simultaneously at different scale no matter whether the anomalies are close to or far from the camera. In fact, the regions marked with yellow grids in Fig.\ref{fig:5} to denote false alarms in accordance with the ground truth can also be true anomalies. For example, in the case shown in Fig.\ref{fig:5}(d), an adult is helping a child pass the turnstile, and for the case shown in Fig.\ref{fig:5}(h), a person is coming up from the turnstile with an irregular jump. Such unusual behaviors are missed in the annotations of the ground truth~\cite{5206569-10} but detected by the proposed method. Besides, it is apparent that the proposed method can accurately detect and localize anomalies in surveillance videos with perspective distortion.

{\bfseries Comparison with the existing approaches}: In Table \ref{Tab3}, we compare quantitatively the proposed method with spatio-temporal compositions~\cite{JAVANROSHTKHARI20131436-16}, dynamic sparse coding (DSC)~\cite{5995524-28}, space-time (ST-MRF)~\cite{5206569-10}, sparse reconstruction cost~\cite{5995434-25}, and local optical flow~\cite{4407716-1} at event level. It can be seen that the proposed method requires the least training data to achieve the same level of high detection rate and low false alarm rate in comparison with spatio-temporal compositions~\cite{JAVANROSHTKHARI20131436-16} and dynamic sparse coding~\cite{5995524-28}.

\begin{table}[!htb]
  \caption{Comparison of abnormal event detection rate and false alarm rate on the Subway dataset}
  \resizebox{0.95\linewidth}{!}{
  \begin{tabular}{|c|cccccc|}
    \hline
    Method & Tp & WD & LT & Misc & Total & FA \\
    \hline
    \hline
    Ground & & {\bfseries 9} & {\bfseries 3} & {\bfseries 7} & {\bfseries19} & {\bfseries 0} \\
    Proposed & 10 & 9 & 3 & 7 & 19 & 2 \\
    STC~\cite{JAVANROSHTKHARI20131436-16} & 15/CL & 9 & 3 & 7 & 19 & 2 \\
    DSC~\cite{5995524-28} & CL & 9 & 3 & 7 & 19 & 2 \\
    ST-MRF~\cite{5206569-10}& 5 & 9 & 3 & 7 & 19 & 3 \\
    SRC~\cite{5995434-25} & 10 & 9 & - & - & 9 & 0 \\
    LOF~\cite{4407716-1} & 5 & 9 & - & - & 9 & 2 \\
	\hline

\multicolumn{7}{l}{TP: Training period; CL: Continuous learning;}\\
\multicolumn{7}{l}{WD: Wrong direction; LT: Loitering; Misc: Miscellany;}\\
\multicolumn{7}{l}{FA: False alarm; “-” means the results are not provided.}\\
  \end{tabular}
  }
  \label{Tab3}
\end{table}

\section{Conclusions}
We associate the optical flows of multiple frames to capture short-term trajectories and introduce a histogram-based shape descriptor, namely, shape contexts, to describe such short-term trajectories, which reflects faithfully the motion trend and details in a local patch. We propose a K-NN similarity-based statistical model to detect anomalies over time and space, which is an unsupervised one-class learning algorithm requiring no clustering nor prior assumption. The anomaly detection through probability can adapt to the whole scene, since the probability is not affected by motion distortions caused by perspective distortion. We carried out experiments on two real-world surveillance videos, for anomaly detection and localization, and the results validate the effectiveness of the proposed method. Moreover, its performance in localizing anomalies outperforms those of the state-of-the-art approaches.

{\small
\bibliographystyle{ieee}
\bibliography{egbib}
}

\end{document}